# Supervised Discriminative Sparse PCA for Com-Characteristic Gene Selection and Tumor Classification on Multiview Biological Data

Chun-Mei Feng, Yong Xu, *Senior Member, IEEE*, Jin-Xing Liu, *Member, IEEE*, Ying-Lian Gao, and Chun-Hou Zheng, *Member, IEEE*

*Abstract*—**Principal component analysis (PCA) has been used to study the pathogenesis of diseases. To enhance the interpretability of classical PCA, various improved PCA methods have been proposed to date. Among these, a typical method is the so-called sparse PCA, which focuses on seeking sparse loadings. However, the performance of these methods is still far from satisfactory due to their limitation of using unsupervised learning methods; moreover, the class ambiguity within the sample is high. To overcome this problem, this paper developed a new PCA method, which is named the supervised discriminative sparse PCA (SDSPCA). The main innovation of this method is the incorporation of discriminative information and sparsity into the PCA model. Specifically, in contrast to the traditional sparse PCA, which imposes sparsity on the loadings, here, sparse components are obtained to represent the data. Furthermore, via the linear transformation, the sparse components approximate the given label information. On the one hand, sparse components improve interpretability over the traditional PCA, while on the other hand, they are have discriminative abilities suitable for classification purposes. A simple algorithm is developed, and its convergence proof is provided. SDSPCA has been applied to the common-characteristic gene selection and tumor classification on multiview biological data. The sparsity and classification performance of SDSPCA are empirically verified via abundant, reasonable, and effective experiments, and the obtained results demonstrate that SDSPCA outperforms other state-of-the-art methods.**

Manuscript received May 7, 2017; revised November 25, 2017, June 24, 2018, and October 8, 2018; accepted January 2, 2019. This work was supported by NSFC under Grant 61872220, Grant 61572284, and Grant 61702299. *(Corresponding authors: Yong Xu; Jin-Xing Liu.)*

C.-M. Feng is with the Bio-Computing Research Center, Harbin Institute of Technology, Shenzhen 518055, China (e-mail: fengchunmei0304@foxmail.com).

Y. Xu is with the Bio-Computing Research Center, Shenzhen Graduate School, Harbin Institute of Technology, Shenzhen 518055, China, and also with the Key Laboratory of Network Oriented Intelligent Computation, Shenzhen 518055, China (e-mail: yongxu@ymail.com).

J.-X. Liu is with the School of Information Science and Engineering, Qufu Normal University, Rizhao 276826, China (e-mail: sdcavell@126.com).

Y.-L. Gao is with the Library of Qufu Normal University, Qufu Normal University, Rizhao 276826, China (e-mail: yinliangao@126.com).

C.-H. Zheng is with the School of Computer Science and Technology, Anhui University, Hefei 230000, China (e-mail: zhengch99@126.com).

This paper has supplementary downloadable material available at http://ieeexplore.ieee.org, provided by the author.

Color versions of one or more of the figures in this paper are available online at http://ieeexplore.ieee.org.

Digital Object Identifier 10.1109/TNNLS.2019.2893190

## I. INTRODUCTION

THE recent development of gene chips enables the simultaneous measure of the expression of thousands of genes [1]. These gene expression data are obtained through the DNA microarray hybridization and next-generation sequencing technology experiments, where the data are typically stored as matrixes [2]–[4]. These techniques yield a data matrix, in which the number of variables $m$ (the gene number) far exceeds the number of samples $n$ [5]. Multiple data sources are typically combined, resulting in the so-called multiview data. Applying a dimensionality reduction method to these multiview data is a reasonable and important step in the biological data processing [6]. "Multiview data" denotes several types of gene expression data from various aspects. Since these data are composed of diverse statistical distributions and semantics, they contain various uncertainties. In general, four different types of biological data are included in multiview data: 1) multiview biological data, measuring samples with categories represented by identical feature (gene) sets; 2) multiview biological data that measure the identical samples with different feature groups; 3) multiview biological data with identical objects measured by the same group of features but under different conditions (expressed as a tensor model: sample × feature × condition); and 4) multiview biological data that measure different features combination with different sample groups under identical conditions [7]. This paper focuses on multiview data obtained from the acquired Cancer Genome Atlas (TCGA; https://tcga-data.nci.nih.gov/) where different groups of samples are represented by the same feature (gene) sets. Here, multiview data are composed of different disease data samples. These samples provide us with an unprecedented opportunity to explore association between different diseases.

One of the most important applications of biological data is the classification of disease samples, e.g., identifying a type of tumor and distinguishing between normal tissue and pathological tissue [8]–[12]. Gene expression data possess the characteristics of high dimension, small sample size, and imbalanced class distribution; however, only a small part of the genes is useful and necessary for the selection of characteristic genes for the tumor classification of samples [8], [13]. The discovered characteristic genes in multiview data can be named common-characteristic (com-characteristic) genes. On the one hand, several genes can be chosen that play an important role for the classification. These genes can improve both the accuracy (ACC) and efficiency of the tumor classification [14]. In the field of disease diagnosis based on the gene expression data, this method can decrease the rate of misdiagnosis [15]. On the other hand, it has been reported that the subset of characteristic genes contains abundant information, which can help to find important disease causing genes and help to explore the potential association between different diseases [16].

The principal component analysis (PCA) is an effective tool to reduce dimensionality and has a wide range of applications throughout biology, face recognition, and classification [8], [17]–[22]. In essence, by minimizing the reconstruction error or by maximizing the data variance, PCA seeks the principal components (PCs) to capture the intrinsically low-dimensional structure that underlies the data. Since the dimensionality of the data has been largely reduced while the latent structure has been preserved, PCA greatly facilitates many applications [23]–[25]. Even though PCA has achieved good results in many areas, several limitations still exist. PCs are obtained via linear combinations of the variables, where the weights (called loadings) are usually all nonzero [26]. Consequently, PCs are the mixtures of a multitude of variables, thus complicating understanding. The data variables often have specific interpretations in different applications. In bioinformatics, a relevant variable might represent a specific gene [27], [28]; in this case, the interpretation of PCs will be greatly facilitated if the loadings have many zero elements. Many researchers have focused on improving the interpretation of PCs. One of the most significant research directions is the investigation of sparse PCA to find sparse loadings.

Sparse techniques have achieved notable effects in many areas, including image annotation and mobile image labeling in the cloud [29]–[31]. Since sparse PCA is widely used, various improved methods on this subject have been presented [32]–[34]. For example, good results on gene expression and ordinary multivariate data have been achieved via Z-SPCA, which was designed using iterative elastic net regression [32]. Semidefinite PCA (DSPCA) proposes that sparse loadings can be found by relaxed semidefinite programing [26]. Shen and Huang [35] obtained sparse loadings through the low-rank matrix factorization, resulting in a method called sPCA-rSVD. GPower has been proposed by Journée *et al.* [36], which identify sparse PCs by solving nonconcave maximization problems; this method includes both a single-unit version and a block-unit version. By adopting expectation maximization in PCA, expectation maximization sparse PCA (EMSPCA) are designed to investigate sparse and nonnegative PCA problems [33]. Based on the augmented Lagrangian method, augmented Lagrangian sparse PCA is proposed to obtain the sparse results in combination with both the interpretable variance and orthogonality [37].

In addition, D'Aspremont *et al.* [34] have investigated a greedy method and found the nonzero elements of sparse loadings; the method is called PathSPCA. Good results have also been obtained by another greedy method, which was designed by Moghaddam *et al.* [38]. Meng *et al.* [39] proposed a coordinate-pair-wise algorithm, which iteratively obtains the pair-wise variables from the sparse PCs on synthetic and gene expression data. Targeting to tensor data, Lai *et al.* [19] extended the sparse method to multilinear PCA. In contrast to methods that impose sparsity constraints on various objective functions, Hu *et al.* [40] designed a method called SPCArt. The motivation of SPCArt was to compute sparse loadings via rotating and truncating the PCA loadings. Unlike sparse PCA methods, which traditionally impose sparsity on the loadings, this paper imposed sparsity on PCs. This leads to a new application of the sparse PCA. The sparse PCs thus obtained enable the physical interpretation of "sparse codes."

However, one important aspect is not covered by either PCA of sparse PCA: the discriminative ability for the classification application. These PCA-based techniques as unsupervised learning methods study training samples without discriminative information and discover the structural knowledge of training samples. This absence of discriminative information may lead to high class ambiguity of the training samples. However, supervised learning methods learn training samples with class labels to predict an unknown data set independent of the training samples. All class labels are known; therefore, the ambiguity of training samples is reduced. PCA as a typical unsupervised learning method does not consider the separation between classes.

This problem has been addressed by the supervised counterpart of PCA: the linear discriminant analysis (LDA), which is a supervised dimensionality reduction method [41]. Various improved supervised methods have been designed. For example, Feng *et al.* [42] joint the projection matrix and discriminative dictionary and obtained supervised dimensionality reduction results. To obtain sparse coding with discriminative ability, linear discriminate analysis and sparseness criterion are jointly worked in SDA [43]. With similar goals, the label consistent K-singular value decomposition has been proposed for face- and object-category recognition [44]. Based on the Fisher discrimination criterion, Yang *et al.* [45] proposed a sparse representation named the Fisher discrimination dictionary learning (FDDL) in the image classification to get discrimination capability. A sparse LDA (SLDA) procedure has been proposed to address high-dimensional small sample size data, and effective results in overfitting and eliminating the data piling problem have been achieved [46]. Lai *et al.* [47] extended LDA to sparse subspace and tensor applications. Unlike supervised methods, which traditionally introduce discriminative information into the model, the N-2-DPCA is designed for the reconstruction of error images and achieves a higher compression rate [48]. N-2-DPCA distinguishes different samples via distance metric, which improves the classification ACC. The nuclear norm is incorporated into this method to compute the structure information. In addition to the sparse PCA method, several PCA-based hybrid models have been proposed recently. For example, to overcome the shortcomings of distributed parameter systems that are difficult to model due to their nonlinearity and infinite dimensionality, a hybrid model is proposed. This model combines decoupled linear autoregressive exogenous and nonlinear radial basis function neural networks via PCA [22]. Moreover, combining PCA with manifold learning yields the probability of representative data points to represent the hash function [49]. This combination has also been applied to biometrics [50], [51].

Although many supervised methods have the ability to discriminate, they cannot capture the characteristic genes well. In bioinformatics, the characteristic genes that can dominate the development of a disease often have large variance due to their own differential expression. The occurrence and development of cancer is mainly caused by a change of these characteristic genes. Therefore, after transforming the data into the $k$ dimension via PCA, characteristic genes can be obtained by the features with the $k$ largest variance. Many supervised discriminative methods focus on the discriminate features of samples and choose the best direction of projection. This, however, does not guarantee that the projection is orthogonal. PCA selects the direction, in which the sample point projection has the largest variance. LDA can generate a subspace with $C - 1$ dimension at most, and thus LDA cannot be used when the dimensionality of the feature vector is larger than $C - 1$ after dimension reduction. The coordinates of the PCA projection are orthogonal, and these new variables (known as PCs) are not related to each other.

Motivated by LDA, Zhang and Li [52] approached the problem under the dictionary learning framework and various supervised methods, where the label matrix was incorporated into the squared-error objective function. The dictionary thus learned to discriminate. It has been reported that the method has achieved good performance with regard to the face recognition. Therefore, it is reasonable to incorporate the label information into PCA and thus achieve discriminative ability. Barshan *et al.* [53] considered the Hibert–Schmidt independence criterion to incorporate the supervised problem into the PCA method. However, unlike the Hibert–Schmidt independence criterion, this paper refers to the successful use of discriminative dictionary learning [52].

This paper proposed a new PCA method, called the supervised discriminative sparse PCA (SDSPCA), which explicitly considers the discriminative information and sparse constraint. This method introduces class labels into the PCA method. Thus, the PCA method is turned into a supervised learning model, which reduces the class ambiguity of the samples. Considering that the sparsity can improve the interpretability of PCs, PCs can combine the supervised label information with sparse constraint to obtain the discriminative ability and interpretability. The proposed method reduces the class ambiguity of the samples, while PCs are expressed in a sparse form. Specifically, sparse components can be obtained to represent the data, while the sparse components approximate the given label information via the linear transformation. To effectively solve the proposed method, an algorithm is provided. To test how well the sparse constraint and the supervised label information work, experiments of the characteristic gene selection and the tumor classification have been conducted. The obtained results show that the proposed method outperforms many of the state-of-the-art PCA methods in the multiview biological data analysis.

The proposed method is formulated in Section II; optimization and convergence analyses are also included. Comprehensive experiments to demonstrate the effectiveness of the proposed method are presented in Section III. Finally, Section IV provides the conclusion with a summary and outlook for future work.

## II. METHODS

In this section, first, a variant formulation of PCA is briefly reviewed. This formulation is consistent with the presented model. Next, the proposed method is introduced. First, several terms and notations are defined.

1) Let $\mathbf{X} = (\mathbf{x}_1, \ldots, \mathbf{x}_n) \in \mathbb{R}^{m \times n}$ be the input matrix, where $n$ represents the number of samples, and $m$ represents the number of variables.
2) Let $\mathrm{Tr}(\mathbf{A})$ be the trace of matrix $\mathbf{A}$.
3) Let $\|\mathbf{Q}\|_{2,1} = \sum_{i=1}^{n} \|\mathbf{q}_i\|_2$ be the $L_{2,1}$ norm, which is obtained by first calculating the $L_2$ norm of each row and then calculating the $L_1$ norm of the resulting $L_2$ norms.
4) Let $\|\mathbf{A}\|_F$ be the Frobenius norm of matrix $\mathbf{A}$.

### A. PCA

Expressed in an alternative way, PCA aims to find a $k$-dimensional $(k \leq m)$ linear subspace to approximate the data matrix as close as possible. Such a subspace was obtained by solving the problem as follows [54]:

$$\min_{\mathbf{Y}, \mathbf{Q}} \|\mathbf{X} - \mathbf{Y}\mathbf{Q}^{\mathrm{T}}\|_F^2 \quad \text{s.t.} \quad \mathbf{Q}^{\mathrm{T}}\mathbf{Q} = \mathbf{I} \tag{1}$$

where $\mathbf{Q} \in \mathbb{R}^{n \times k}$ and $\mathbf{Y} = (\mathbf{y}_1, \ldots, \mathbf{y}_k) \in \mathbb{R}^{m \times k}$. Each row of $\mathbf{Q}$ represents the projection of a data point in the subspace; its solution corresponds to normalized PCs. The normality is caused by the constraint $\mathbf{Q}^{\mathrm{T}}\mathbf{Q} = \mathbf{I}$. Each column of $\mathbf{Y}$ represents a principal direction; its solution corresponds to the scaled loading. If orthonormality is imposed on $\mathbf{Y}$ instead, classical PCA can be recovered: $\mathbf{Y}$ corresponds to the loadings, while $\mathbf{Q}$ are PCs.

### B. Supervised Discriminative Sparse PCA

SDSPCA obtains PCs by incorporating supervised label information and sparse constraints. The proposed method overcomes the shortcomings of classical PCA where PCs are dense; in particular, it is theoretically demonstrated that the class ambiguity of the sample is greatly reduced. The following optimization is formulated to realize SDSPCA:

$$\min_{\mathbf{Y}, \mathbf{Q}, \mathbf{A}} \|\mathbf{X} - \mathbf{Y}\mathbf{Q}^{\mathrm{T}}\|_F^2 + \alpha \|\mathbf{B} - \mathbf{A}\mathbf{Q}^{\mathrm{T}}\|_F^2 + \beta \|\mathbf{Q}\|_{2,1}$$
$$\text{s.t.} \quad \mathbf{Q}^{\mathrm{T}}\mathbf{Q} = \mathbf{I} \tag{2}$$

where $\alpha$ and $\beta$ are the scale weights. $\mathbf{B} \in \mathbb{R}^{c \times n}$ represents the class indicator matrix. The class indicator matrix consists of elements 0 and 1. The position of element 1 in each column represents the class labels. This matrix can be defined as

$$\mathbf{B}_{i,j} = \begin{cases} 1, & \text{if } s_j = i, \ j = 1, 2\ldots, n, \ i = 1, 2, \ldots, c \\ 0, & \text{otherwise} \end{cases} \tag{3}$$

where $c$ represents the number of classes in training data and $s_j \in \{1, \ldots, c\}$ represents the class label. For example, assuming that the label of each data point gnd $= [1, 2, 4, 3, 3, 1, 5, 6]^{\mathrm{T}}$, $\mathbf{B}$ can be defined as follows:

$$\mathbf{B} = \begin{bmatrix} 1 & 0 & 0 & 0 & 0 & 1 & 0 & 0 \\ 0 & 1 & 0 & 0 & 0 & 0 & 0 & 0 \\ 0 & 0 & 0 & 1 & 1 & 0 & 0 & 0 \\ 0 & 0 & 1 & 0 & 0 & 0 & 0 & 0 \\ 0 & 0 & 0 & 0 & 0 & 0 & 1 & 0 \\ 0 & 0 & 0 & 0 & 0 & 0 & 0 & 1 \end{bmatrix}. \tag{4}$$

Matrix $\mathbf{A} \in \mathbb{R}^{e \times k}$ was arbitrarily initialized as a linear transformation matrix to easily obtain the solution.

SDSPCA belongs to a type of novel supervised PCA methods, which explicitly incorporates both label information and sparsity into PCA. The second and third terms in the objective function guarantee that SDSPCA can gain the discriminative ability and interpretability, respectively. $L_{2,1}$ norm makes $Q$ robust to outliers, while inducing sparsity [55]–[57]. Thus, the data can be represented with PCs having both discriminative power and sparsity.

### C. Optimization

This section introduces the detailed optimization process of SDSPCA. Since the closed-form solution of SDSPCA is not available, following a previous study, an alternative solution is provided [55]. $\mathbf{Y}$, $\mathbf{Q}$, and $\mathbf{A}$ are updated alternatively. For convenience, $\ell = \|\mathbf{X} - \mathbf{Y}\mathbf{Q}^{\mathrm{T}}\|_F^2 + \alpha\|\mathbf{B} - \mathbf{A}\mathbf{Q}^{\mathrm{T}}\|_F^2 + \beta\|\mathbf{Q}\|_{2,1}$. Note that $\|\mathbf{Q}\|_{2,1}$ is convex; however, its derivative does not exist when $\mathbf{q}_i = 0$. When $\mathbf{q}_i \neq 0$, the optimal $\mathbf{Y}$ can be solved, while fixing the others. The derivative is

$$\frac{\partial \ell(\mathbf{Y})}{\partial \mathbf{Y}} = -2(\mathbf{X} - \mathbf{Y}\mathbf{Q}^{\mathrm{T}})\mathbf{Q}. \tag{5}$$

Making the derivative equal to 0 yields

$$\mathbf{Y} = \mathbf{X}\mathbf{Q}. \tag{6}$$

Similarly, for $\mathbf{A}$

$$\mathbf{A} = \mathbf{B}\mathbf{Q}. \tag{7}$$

The solution of $\mathbf{Q}$ is more complicated and will be approached by applying a technique of $L_{2,1}$ norm optimization. An auxiliary objective function will be used and (2) yields

$$\min_{\mathbf{Y}, \mathbf{Q}, \mathbf{A}, \mathbf{V}} \mathrm{Tr}(\mathbf{X} - \mathbf{Y}\mathbf{Q}^{\mathrm{T}})(\mathbf{X} - \mathbf{Y}\mathbf{Q}^{\mathrm{T}})^{\mathrm{T}}$$
$$+ \alpha\mathrm{Tr}(\mathbf{B} - \mathbf{A}\mathbf{Q}^{\mathrm{T}})(\mathbf{B} - \mathbf{A}\mathbf{Q}^{\mathrm{T}})^{\mathrm{T}} + \beta\mathrm{Tr}(\mathbf{Q}^{\mathrm{T}}\mathbf{V}\mathbf{Q}) \tag{8}$$

where $\mathbf{V} \in \mathbb{R}^{n \times n}$ represents a diagonal matrix, whose element is expressed as

$$\mathbf{V}_{ii} = \frac{1}{2\|\mathbf{q}_i\|_2}. \tag{9}$$

Substituting the solutions of $\mathbf{Y}$ and $\mathbf{A}$ into (8), we have

$$\mathrm{Tr}(\mathbf{X} - \mathbf{X}\mathbf{Q}\mathbf{Q}^{\mathrm{T}})(\mathbf{X} - \mathbf{X}\mathbf{Q}\mathbf{Q}^{\mathrm{T}})^{\mathrm{T}}$$
$$+ \alpha\mathrm{Tr}(\mathbf{B} - \mathbf{B}\mathbf{Q}\mathbf{Q}^{\mathrm{T}})(\mathbf{B} - \mathbf{B}\mathbf{Q}\mathbf{Q}^{\mathrm{T}})^{\mathrm{T}} + \beta\mathrm{Tr}(\mathbf{Q}^{\mathrm{T}}\mathbf{V}\mathbf{Q})$$
$$= -\mathrm{Tr}(\mathbf{Q}^{\mathrm{T}}\mathbf{X}^{\mathrm{T}}\mathbf{X}\mathbf{Q}) + \|\mathbf{X}\|_F^2 - \alpha\mathrm{Tr}(\mathbf{Q}^{\mathrm{T}}\mathbf{B}^{\mathrm{T}}\mathbf{B}\mathbf{Q})$$
$$+ \alpha\|\mathbf{B}\|_F^2 + \beta\mathrm{Tr}(\mathbf{Q}^{\mathrm{T}}\mathbf{V}\mathbf{Q})$$
$$= \mathrm{Tr}(\mathbf{Q}^{\mathrm{T}}(-\mathbf{X}^{\mathrm{T}}\mathbf{X} - \alpha\mathbf{B}^{\mathrm{T}}\mathbf{B} + \beta\mathbf{V})\mathbf{Q}) + \|\mathbf{X}\|_F^2 + \alpha\|\mathbf{B}\|_F^2. \tag{10}$$

Therefore, (10) is equivalent to the following problem [54]:

$$\ell(\mathbf{Q}) = \min_{\mathbf{Q}^{\mathrm{T}}\mathbf{Q} = \mathbf{I}} \mathrm{Tr}\mathbf{Q}^{\mathrm{T}}(-\mathbf{X}^{\mathrm{T}}\mathbf{X} - \alpha\mathbf{B}^{\mathrm{T}}\mathbf{B} + \beta\mathbf{V})\mathbf{Q}. \tag{11}$$

Optimal $\mathbf{Q}$ can be solved by computing the eigenvectors of the matrix $\mathbf{Z} = -\mathbf{X}^{\mathrm{T}}\mathbf{X} - \alpha\mathbf{B}^{\mathrm{T}}\mathbf{B} + \beta\mathbf{V}$. Eigenvectors are picked corresponding to the $k$ smallest eigenvalues.

The following explains why a sparse solution can be obtained by minimizing (11) [58]. According to the definition of $\mathbf{V}_{ii}$, the solution of $\mathbf{Q}$ is naturally sparse. Specifically, if $\|\mathbf{q}_i\|_2$ is small, then $\mathbf{V}_{ii}$ will be large and (9) tends to obtain $\mathbf{q}_i$ with a much smaller $L_2$ norm. With the increasing number of iterations, the norms of $\mathbf{q}_i$ are close to zero and a sparse $\mathbf{Q}$ can be obtained. $\mathbf{Q}$, $\mathbf{Y}$, and $\mathbf{A}$ can be iteratively updated until the objective value stabilizes. The detailed procedure is presented in Algorithm 1.

### D. Convergence Analysis

As shown in Algorithm 1, SDSPCA follows an alternative process. The objective function in (2) is nonincreasing against the updating steps. By Lemma 1, the optimization procedure can be demonstrated to be monotonically decreasing.

---

**Algorithm 1** The Algorithm of Supervised Discriminative Sparse PCA (SDSPCA)

---

**Input**: Data matrix $\mathbf{X} = (\mathbf{x}_1, \ldots, \mathbf{x}_n) \in \mathbb{R}^{m \times n}$,
　indicator matrix $\mathbf{B} \in \mathbb{R}^{c \times n}$ and parameters $\alpha$, $\beta$
**Output**: $\mathbf{Y}$, $\mathbf{Q}$

---

**Initialize:** $\mathbf{V} = \mathbf{I}_{n \times n}$, randomly initialize matrices $\mathbf{A} \in \mathbb{R}_c^{c \times k}$;
**Repeat**:
　1. Update $\mathbf{Q}$ by solving the problem in Eq.(11);
　2. Update $\mathbf{Y}$ by (6);
　3. Update $\mathbf{A}$ by (7);
　4. Update $\mathbf{V}$ by (9);
**Until** Convergence

---

*Lemma 1:* For any nonzero vectors $\mathbf{a}, \mathbf{a}_0 \in \mathbb{R}^m$, the following result can be obtained [55]:

$$\|\mathbf{a}\|_2 - \frac{\|\mathbf{a}\|_2^2}{2\|\mathbf{a}_0\|_2} \leq \|\mathbf{a}_0\|_2 - \frac{\|\mathbf{a}_0\|_2^2}{2\|\mathbf{a}_0\|_2}. \tag{12}$$

The convergence behavior of SDSPCA can be shown as follows.

*Proof:* According to Algorithm 1, the following inequality results:

$$
\begin{aligned}
&\|\mathbf{X} - \mathbf{Y}^{t+1}(\mathbf{Q}^{t+1})^{\mathrm{T}}\|_F^2 + \alpha\|\mathbf{B} - \mathbf{A}^{t+1}(\mathbf{Q}^{t+1})^{\mathrm{T}}\|_F^2 \\
&\quad + \beta tr((\mathbf{Q}^{t+1})^{\mathrm{T}}\mathbf{V}^t\mathbf{Q}^{t+1}) \\
&\leq \|\mathbf{X} - \mathbf{Y}^t(\mathbf{Q}^t)^{\mathrm{T}}\|_F^2 + \alpha\|\mathbf{B} - \mathbf{A}^t(\mathbf{Q}^t)^{\mathrm{T}}\|_F^2 \\
&\quad + \beta tr((\mathbf{Q}^t)^{\mathrm{T}}\mathbf{V}^t\mathbf{Q}^t).
\end{aligned}
\tag{13}
$$

Since $\|\mathbf{Q}\|_{2,1} = \sum_{i=1}^{n}\|\mathbf{q}_i\|_2$, the above inequality can be rewritten as

$$
\begin{aligned}
&\|\mathbf{X} - \mathbf{Y}^{t+1}(\mathbf{Q}^{t+1})^{\mathrm{T}}\|_F^2 + \alpha\|\mathbf{B} - \mathbf{A}^{t+1}(\mathbf{Q}^{t+1})^{\mathrm{T}}\|_F^2 \\
&\quad + \beta\sum_{i=1}^{n}\left(\frac{\|\mathbf{q}_i^{t+1}\|_2^2}{2\|\mathbf{q}_i^t\|_2} - \|\mathbf{q}_i^{t+1}\|_2\right) \\
&\leq \|\mathbf{X} - \mathbf{Y}^t(\mathbf{Q}^t)^{\mathrm{T}}\|_F^2 + \alpha\|\mathbf{B} - \mathbf{A}^t(\mathbf{Q}^t)^{\mathrm{T}}\|_F^2 \\
&\quad + \beta\sum_{i=1}^{n}\left(\frac{\|\mathbf{q}_i^t\|_2^2}{2\|\mathbf{q}_i^t\|_2} - \|\mathbf{q}_i^{t+1}\|_2\right).
\end{aligned}
\tag{14}
$$

Recalling Lemma 1

$$
\frac{\|\mathbf{q}_i^{t+1}\|_2^2}{2\|\mathbf{q}_i^t\|_2} - \|\mathbf{q}_i^{t+1}\|_2 \geq \frac{\|\mathbf{q}_i^t\|_2^2}{2\|\mathbf{q}_i^t\|_2} - \|\mathbf{q}_i^t\|_2.
\tag{15}
$$

Combing the above two inequalities yields the following result:

$$
\begin{aligned}
&\|\mathbf{X} - \mathbf{Y}^{t+1}(\mathbf{Q}^{t+1})^{\mathrm{T}}\|_F^2 + \alpha\|\mathbf{B} - \mathbf{A}^{t+1}(\mathbf{Q}^{t+1})^{\mathrm{T}}\|_F^2 + \beta\|\mathbf{Q}^{t+1}\|_{2,1} \\
&\leq \|\mathbf{X} - \mathbf{Y}^t(\mathbf{Q}^t)^{\mathrm{T}}\|_F^2 + \alpha\|\mathbf{B} - \mathbf{A}^t(\mathbf{Q}^t)^{\mathrm{T}}\|_F^2 + \beta\|\mathbf{Q}^t\|_{2,1}.
\end{aligned}
\tag{16}
$$

Thus, the algorithm will decrease the objective value in each iteration.

## III. EXPERIMENTS

To evaluate the performances, especially the sparsity and ambiguity in training samples, the proposed method is applied to different situations, such as com-characteristic gene selection and tumor classification. For comparison, supervised learning methods such as LDA [41], SLDA [46], and SDA [43] as well as unsupervised learning methods such as the classical PCA [17] and five state-of-the-art sparse PCA methods have also been tested. These sparse PCA methods include EMSPCA [33], Z-SPCA [32], PathSPCA [34], SPCArt [40], and N-2-DPCA [48]. Since EMSPCA, Z-SPCA, PathSPCA, and SPCArt are currently the most advanced sparse PCA methods, and LDA, SLDA, and SDA are classical supervised dimensionality reduction methods, the sparsity and ability of disentangling sample ambiguity of the proposed method can be reasonably proved. The detailed information of the used data sets and the experimental results both in the com-characteristic gene selection and the tumor classification are summarized in the following.

### A. Data Sets

Multiview data of gene expression have been used to evaluate the performance of the proposed method: including pancreatic cancer (PAAD), head and neck squamous cell carcinoma (HNSC), and cholangiocarcinoma (CHOL) gene expression data. These data are composed of different groups of samples represented by the same feature (gene) sets and can be downloaded from TCGA. All samples in the multiview data are randomly divided into training and testing sets. Since the normal samples of each disease data originate from different tissues, the normal and disease data were classified into four categories. Table I presents a summary of the multiview data with four classes. Each data set consists of different groups of samples represented by the same feature (gene) sets.

### B. Exploration of SDSPCA Properties

The properties of SDSPCA were explored, including the parameters of different scale weights on the classification ACC and the computational complexities. Fig. 1 shows the influence of different scale weights of parameters from $\{10^{-20}, \ldots, 10^{20}\}$ for the classification ACC. Fig. 1 shows that the SDSPCA can achieve best performance within $\alpha \in \{10^{-20}, \ldots, 10^{15}\}$ and $\beta \in \{10^{-20}, \ldots, 10^{-5}\}$. The classification results in this range are stable and optimal. The high value of $\beta$ will cause data loss due to the high sparsity, and the classification ACC is not satisfied.

Here, an analysis of the computational complexities is presented for different methods. Since the eigenvalue decomposition is the most time-consuming step, the computational complexity is $O(D^3)$. $D$ represents the dimensionality of

TABLE I
SUMMARY OF THE MULTIVIEW DATA WITH FOUR CLASSES

| Datasets | Classes | Number of | |
| --- | --- | --- | --- |
| | | Samples | Genes |
| PAAD | 1 | 176 | 20,502 |
| HNSC | 2 | 398 | 20,502 |
| CHOL | 3 | 36 | 20,502 |
| Normal | 4 | 33 | 20,502 |
| multi-view data | | 643 | 20,502 |

Notes: The first three categories are diseased samples of different data sets.

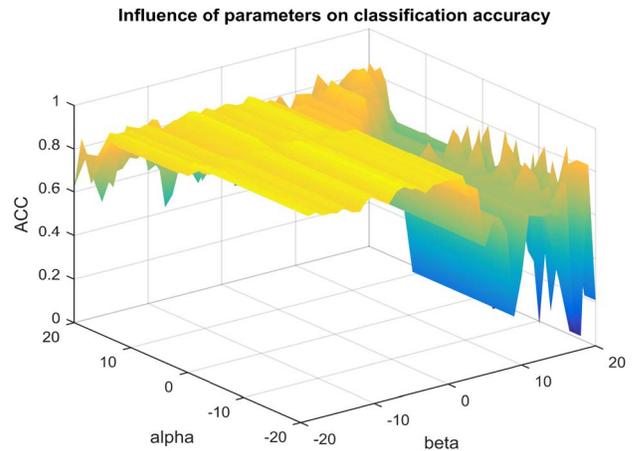

Fig. 1. Influence of different scale weights of parameters on the classification ACC. The three coordinates represent the classification ACC and the exponent values of parameters $\alpha$ and $\beta$, respectively.

TABLE II
COMPUTATION TIME OF EACH METHOD

| Methods | Time (s.) |
|---|---|
| LDA | 0.2357±5.6217e-04 |
| SLDA | 164.1045±0.7004 |
| SDA | 11.7580±0.0116 |
| PCA | 0.1863±0.1843e-05 |
| EMSPCA | 0.3593±2.6414e-04 |
| N-2-DPCA | 208.7232±17.7250 |
| Z-SPCA | 2.3938±9.6255e-04 |
| PathSPCA | 0.1535±7.0102e-05 |
| SPCArt | 1.8408±9.2656e-04 |
| SDSPCA | 0.5264±0.0101e-05 |

Notes: SLDA and SDA have been preprocessed with PCA.

the original data. In this experiment, the experimental data have the characteristics of high dimension and small sample. Thus, the computational complexity of the utilized method is $O(z(mn^2))$, where $z$ represents the number of iterations. The computational complexity of SDSPCA increases linearly with the data dimension. The dimension is the decisive factor that determines the computational complexity. Traditional PCA and LDA have the computational complexity $O(mn^2)$. SLDA and SDA first decompose the eigenvalue, and then solve both LASSO and elastic net problems. Their computational complexity is $O(d(mn^2))$, where $d$ represents the iteration number in LASSO or elastic net problem [47]. The complexity of SPCArt is $zO(k^2m + k^3)$, and the overhead of preprocessing and initialization is ignored (SPCArt has the additional cost of PCA) [40]. Since Z-SPCA solves the problem via the iterative regression, its computational complexity is $O(z(k^2m + knm))$. PathSPCA as a combinatorial greedy algorithm and its computational complexity are $O(kcnm + kc^3)$, where $c$ represents the cardinality of a loading [40]. EMSPCA presents a sparse PCA algorithm with the computational complexity $O(mn)$ [33]. However, different implementations of the same method may require different times. The following summarizes the numerical results.

To investigate how fast the proposed algorithm converges, the execution times presented in Table II with $k = 1$ were used to specify the size of sparse components. Every method is repeated ten times, and the average and variance values are calculated. These experiments are executed on a PC with Intel core i7-6700 CPU and 8-GB RAM. In Table II, N-2-DPCA takes the longest time to finish the loop. PCA and LDA are basically equally time-consuming. Since other methods are iterative algorithms and linearly related to the dimension of the data, their time cost is much larger. Since LASSO and elastic net cost significant time in the iteration process of high-dimensional data, both SLDA and SDA have been preprocessed with PCA. Soft thresholding is used in Z-SPCA to calculate the components, which reduces the time cost. However, the speed of SDSPCA is stable and still within PCA.

## C. Experimental Settings

LDA, SLDA, SDA, PCA, EMSPCA, N-2-DPCA, Z-SPCA, PathSPCA, SPCArt, and SDSPCA are used for both the gene selection and the tumor classification. The multiview biological data are divided into training and testing sets. For all the unsupervised learning methods involved in this experiment, $k$ represents the number of sparse components to obtain the desired dimensional data. In the experiment of the com-characteristic gene selection, $k = 1$ to specify the size of sparse components. In the tumor classification experiment, the values of $k$ from 1 to 50 are recorded. Since the multiview data have an imbalanced number of samples between different classes, data imbalances were either eliminated or reduced by changing the distribution of training samples. In the experiment, part of the data was extracted first (143 samples, which is about one-sixth) and reserved for the final test. Other data are randomly divided as training and validation data for fivefold cross-validation. All data are randomly extracted, and the imbalanced data distribution can be obtained by averaging the results. For supervised learning method LDA, assuming that the number of categories is $C$, the data can be reduced to $1 \sim C - 1$ dimension subspaces. In this experiment, the dimensions of LDA are 1–3. Since the dimension of the data set is too high, SLDA and SDA were preprocessed with PCA to reduce the required computation time. For the SDSPCA method, parameters $\alpha$ and $\beta$ are selected from $\{10^{-20}, \ldots, 10^{20}\}$. Other parameters in several methods need to be set in advance. For the Z-SPCA method, $\lambda$ was set to infinity. Soft thresholding is conducted to calculate the components for the biological data of high dimensional and small sample. For the SPCArt method, $\lambda^* = 1/\sqrt{m}$ was set to control the sparsity. This value not only guarantees the sparsity but also avoids the truncation of a zero vector.

## D. Results of Com-Characteristic Gene Selection

A novel approach in bioinformatics aims to conduct the gene selection in machine learning, instead of selecting genes from the original data. The discovery of biological processes by machine learning is biologically significant for the classification required for disease prediction. Furthermore, characteristic genes selected in these processes help to improve ACC of the classification performance. Biological data are usually noisy, which is closely associated with the performance of disease prediction and tumor classification. The sparse constraint in the proposed method greatly improves the interpretability of the classical PCA method. After the sparse constraint of biological data, the noise and the genes that are not relevant in the course of the disease have been ignored. The marker genes can be found by the characteristic gene selection for disease prediction and reveal the complex mechanism of the underlying pathway. It is necessary to test how well sparse coding helps in the characteristic gene selection, which the proposed method proved reasonably. The discovered characteristic genes of the multiview data can be named com-characteristic genes. To discover the mechanism of action of com-characteristic genes and to explore the potential connection between different diseases (in multiview biological data) for early diagnosis and treatment of cancer, SDSPCA is applied to highlight com-characteristic genes.

*1) Observations and Discussions of Com-Characteristic Genes:* For the comparison, 500 genes were selected by

TABLE III
Results on Multiview Data of LDA, SLDA, SDA, PCA, EMSPCA, N-2-DPCA, Z-SPCA, PathSPCA, SPCArt, and SDSPCA

| Methods | Com-characteristic genes | No. |
|---|---|---|
| LDA | **GNAS**,TIMP2,**KRT8**,THBS1,HIF1A,ITGB1,MMP14,BSG,PRDX1,MMP7,TTR,FGA,MMP2,BCL2L1,PLAU,NQO1,CXCL12,GPC3,TKT,MUC2,CDH1,VEGFA,CFTR,ERBB3,NOTCH1,SERPINB5,MME,KRT20,SLC2A1,LGALS3,SOX2,TP63,CA9,HPGD,HP,ALCAM,MCM7,AQP1 | 38 |
| SLDA | SST,**GNAS**,CEACAM5,VIM,HLA-A,COL1A1,KRT7,KRT19,KRT18,TIMP2,PKM,ALDOA,ENO1,MMP11,**KRT8**,THBS1,TIMP1,LDHA,HIF1A,CD44,GSTP1,TGB1,TIMP3,HSPG2,PRDX1,DUSP1,MMP7,SERPINA3,FGA,MMP2,MUC1,SPP1,CTRL,EPCAM,TYMP,ALB,LAMC2,CFTR,SERPINB5,CYP2E1,SLC2A1,TP63,SFN,S100A9,FSCN1,ITGA6,SLC7A5,AQP1 | 49 |
| SDA | SST,**GNAS**,CEACAM5,CTSB,VIM,HLA-A,COL1A1,KRT7,KRT19,KRT18,TIMP2,PKM,ALDOA,ENO1,MMP11,**KRT8**,THBS1,TIMP1,LDHA,HIF1A,CD44,GSTP1,ITGB1,TIMP3,HSPA5,PRDX1,MMP7,SERPINA3,FGA,MMP2,MUC1,SPP1,EGFR,EPCAM,TYMP,PLAU,ALB,LAMC2,MUC2,TNFSF10,CFTR,SERPINB5,CYP2E1,ANXA1,SLC2A1,TP63,SFN,S100A9,FSCN1,ITGA6,SLC7A5,S100A2,CD151,AQP1 | 54 |
| PCA | **GNAS**,CTSB,VIM,HLA-A,COL1A1,KRT19,ANXA2,PKM,ALDOA,CTNNB1,STAT3,ENO1,**KRT8**,THBS1,LDHA,HIF1A,CD44,GSTP1,MCL1,ITGB1,MMP14,HSPA5,BSG,HSPG2,DCN,PRDX1,DUSP1,SERPINA3,TTR,FGA,MMP2,SPP1,CCND1,TYMP,ALB,TKT,LAMC2,CDH1,SERPINB5,ANXA1,SLC2A1,CTNNA1,TP63,SFN,S100A9,FSCN1,ITGA6,HP,LGALS1,SLC7A5,S100A2,POSTN,CTSC | 53 |
| EMPCA | **GNAS**,CTSB,VIM,HLA-A,COL1A1,KRT19,ANXA2,PKM,ALDOA,CTNNB1,STAT3,ENO1,**KRT8**,THBS1,LDHA,HIF1A,CD44,GSTP1,MCL1,ITGB1,MMP14,HSPA5,BSG,HSPG2,DCN,PRDX1,DUSP1,MX1,FGA,MMP2,CCND1,TYMP,PLAU,ALB,TKT,LAMC2,CDH1,SERPINB5,ANXA1,SLC2A1,CTNNA1,TP63,SFN,S100A9,FSCN1,ITGA6,HP,LGALS1,SLC7A5,S100A2,POSTN,HMGA1,CTSC | 53 |
| N-2-DPCA | **GNAS**,ALB,S100A9,COL1A1,SFN,PKM,ENO1,HP,ANXA1,HLA-A,S100A2,ALDOA,**KRT8**,CTSB,LAMC2,LDHA,ANXA2,GSTP1,SLC2A1,FGA,CDH1,CD44,ITGA6,HSPA5,FSCN1,VIM,SLC7A5,ITGB1,MMP14,KRT19,BSG,SERPINB5,MCL1,TP63,PRDX1,HSPG2,TTR,TYMP,SERPINA3,CTNNA1,THBS1,MMP2,DUSP1,CTSC,HIF1A,LGALS1,DCN,CTNNB1,TKT,CCND1,SPP1,POSTN,STAT3 | 53 |
| Z-SPCA | **GNAS**,CTSB,VIM,HLA-A,COL1A1,KRT19,ANXA2,PKM,ALDOA,CTNNB1,STAT3,ENO1,**KRT8**,THBS1,LDHA,HIF1A,CD44,GSTP1,MCL1,ITGB1,MMP14,HSPA5,BSG,HSPG2,DCN,PRDX1,DUSP1,SERPINA3,TTR,FGA,MMP2,SPP1,CCND1,TYMP,ALB,TKT,LAMC2,CDH1,SERPINB5,ANXA1,SLC2A1,CTNNA1,TP63,SFN,S100A9,FSCN1,ITGA6,HP,LGALS1,SLC7A5,S100A2,POSTN,CTSC | 53 |
| PathSPCA | **GNAS**,CTSB,VIM,HLA-A,COL1A1,KRT7,KRT19,ANXA2,KRT18,PKM,ALDOA,ENO1,**KRT8**,TIMP1,LDHA,CD44,GSTP1,MCL1,HSPA5,RHOC,BSG,MMP7,SERPINA3,MMP2,MUC1,SPP1,CTRL,EPCAM,ALB,TKT,CDH1,CFTR,LGALS3,PDCD4,HMGA1,CD151,AQP1,PPIA | 39 |
| SPCArt | **GNAS**,CTSB,VIM,HLA-A,COL1A1,KRT19,ANXA2,KRT18,PKM,ALDOA,ENO1,**KRT8**,THBS1,LDHA,HIF1A,CD44,GSTP1,MCL1,ITGB1,MMP14,HSPA5,BSG,HSPG2,PRDX1,DUSP1,SERPINA3,TTR,FGA,MMP2,TYMP,ALB,TKT,F2,LAMC2,CDH1,SERPINB5,CYP2E1,ANXA1,SLC2A1,CTNNA1,TP63,SFN,CYP2A6,S100A9,FSCN1,ITGA6,HP,LGALS1,SLC7A5,S100A2,POSTN,KNG1,CTSC | 53 |
| SDSPCA | **GNAS**,CEACAM5,CTSB,VIM,COL1A1,KRT7,KRT19,ANXA2,KRT18,TIMP2,PKM,ALDOA,ENO1,MMP11,**KRT8**,THBS1,TIMP1,LDHA,CD44,GSTP1,ANXA5,ITGB1,MMP14,RHOC,HSPG2,PRDX1,MMP7,SERPINA3,TTR,FGA,SPP1,CCND1,***PDGFRB***,EPCAM,BCL2L1,PLAU,ALB,NQO1,***ABCC1***,LAMC2,CDH1,VEGFA,CFTR,***MUC4***,ERBB3,CYP2E1,ANXA1,***FGFR4***,TP63,CYP2A6,S100A9,ITGA6,HP,S100A2,***YAP1***,POSTN,CD151,AQP1,***MB***,CTSC | 60 |

Notes: Com-characteristic genes are selected by the listed methods. Underlined genes denote the same Com-characteristic genes found by all methods. The genes in italic are the Com-characteristic genes which our method can select, while the other methods cannot. "No." denotes the number of selected Com-characteristic gene in each method.

TABLE IV
Summary Function of the Same Com-Characteristic Genes Found by All Methods

| Gene ID | Gene | Summary of function | Relevance score |
|---|---|---|---|
| 2778 | GNAS | The GNAS locus is imprinted in a complex manner, suggesting distinct paternally, maternally and biallelically expressed proteins. | 24.49, 6.23, 1.54 |
| 3856 | KRT8 | This gene is a member of the type II keratin family clustered on the long arm of chromosome 12. | 7.88, 18.13, 4.98 |

Notes: The relevance score indicates the degree of a gene associated with a disease. Here, these diseases are referred to as PAAD, HNSC, and CHOL.

each compared method. The common virulence gene pool of PAAD, HNSC, and CHOL was downloaded from GeneCards. GeneCards is a searchable, integrative database that provides large data sets on annotated and predicted human genes. The public website is http://www.genecards.org/.

The selected genes from each method were matched to the downloaded common virulence gene pool to find the com-characteristic genes. Table III lists the matching results on multiview data with respect to eight methods. The complete results of the common virulence genes can be found in the supplementary file. Com-characteristic genes from each method were selected and underlined genes denote the same com-characteristic genes found by these methods. Genes in italics indicate the com-characteristic genes, and the proposed method can select but the other methods cannot. Table III shows that the proposed method offers an advantage over the other tested methods on multiview data. The proposed method has the highest identification number compared to other methods. Since the sparse constraint was introduced into PCA, interpretation of the computed PCs will be greatly facilitated. GNAS and KRT8 are both found by all compared methods. PDGFRB, ABCC1, MUC4, FGFR4, YAP1, and MB are com-characteristic genes, which can be selected by the proposed method, while other methods cannot. These are the genes that cannot be ignored in the study of the relationship among the three diseases. The detailed information of these genes is summarized in the following.

Table IV summarizes the functions of the underlined genes, which are the same com-characteristic genes found by all the methods. GNAS produces a series of biological products via many promoters [59]. The lesions of GNAS cause Mosaic, Somatic, Mccune-Albright Syndrome,

TABLE V

Summary Function of Com-Characteristic Genes Which Our Method Can Select, While the Other Methods Cannot. Relevance Score Indicates the Degree of a Gene Associating With a Disease. These Diseases Are Referred to as PAAD, HNSC, and CHOL

| Gene ID | Gene | Summary of function | Relevance score |
|---|---|---|---|
| 5159 | PDGFRB | This gene encodes a cell surface tyrosine kinase receptor for members of the platelet-derived growth factor family. | 39.55, 35.73, 1.09 |
| 4363 | ABCC1 | The protein encoded by this gene is a member of the superfamily of ATP-binding cassette (ABC) transporters. ABC proteins transport various molecules across extra-and intra-cellular membranes. | 17.36, 18.72, 1.09 |
| 4585 | MUC4 | This gene may play a role in tumor progression. | 29.89, 16.54, 9.84 |
| 2264 | FGFR4 | The protein encoded by this gene is a member of the fibroblast growth factor receptor family, where amino acid sequence is highly conserved between members and throughout evolution. | 17.4, 14.78, 1.09 |
| 10412 | YAP1 | This gene is known to play a role in the development and progression of multiple cancers as a transcriptional regulator of this signaling pathway and may function as a potential target for cancer treatment. | 7.36, 10.85, 1.54 |
| 4151 | MB | This gene encodes a member of the globin superfamily and is expressed in skeletal and cardiac muscles. | 3.13, 6.98, 1.09 |

Notes：Relevance score indicates the degree of a gene associating with a disease. Here, these diseases are referred to as PAAD, HNSC, CHOL.

and PseudohypoparathyroidismIc. The relevance scores of KRT8 with the three diseases show their close relationship. However, few medical studies on KRT8 and these diseases have been published. KRT8 can be used as a newly discovered com-characteristic gene for the exploration of the potential connection among these different diseases. The relevance scores of GNAS with PAAD, HNSC, and CHOL are 24.49, 6.23, and 1.54, respectively. The relevance scores of KRT8 with these three diseases are 7.88, 18.13, and 4.98, respectively. Both of these genes have a high correlation score with the corresponding disease. These genes cannot be ignored when studying the links among the three diseases in these multiview biological data.

To highlight the effectiveness of the proposed method, it is necessary to analyze the functions of the obtained com-characteristic genes that have not been found by other methods. Table V summarizes the functions and relevance scores for each disease associated with these genes. Each gene has a high correlation with disease, especially with PAAD. PDGFRB is a critical factor of p53 (mut)-driven metastas is in PAAD [60]. More medical studies should focus on the relationship between PDGFRB and HNSC, since this gene has a high relevance score with HNSC. ABCC1 plays a multidrug transporter role in cancer, and the high expression of this gene indicates a warning in CHOL [61], [62]. FGFR4 and MUC4 have been studied before, as well as the relationship among these three diseases; therefore, these genes cannot be ignored when studying the relationship among the three diseases [63]–[67]. YAP1 activates oncogene Kras and increases the risk of PAAD [68]. These genes indicate that the presented method can obtain more com-characteristic genes. These genes can prove that a specific relationship exists among the three diseases, providing an unprecedented opportunity to explore the potential connection among different diseases. New cancer markers can be studied from these com-characteristic genes, and the potential connection among different diseases can be explored for the early diagnosis and treatment of cancer. Further medical research should focus on these genes to discover new potential relationships with these three diseases.

TABLE VI

Summary of the Pathways Found by Our Method. Ten Pathways Are Included in This Table and in Ascending Order by the *P*-Value

| Gene Set Name | Genes in Overlap | p-value | FDR q-value |
|---|---|---|---|
| ECM-receptor interaction | 8 | 1.82 E-13 | 3.38 E-11 |
| Focal adhesion | 9 | 5.09 E-12 | 4.73 E-10 |
| Pathways in cancer | 9 | 3.96 E-10 | 2.45 E-8 |
| Small cell lung cancer | 5 | 8.38 E-8 | 3.89 E-6 |
| Bladder cancer | 4 | 2.64 E-7 | 9.83 E-6 |
| Pathogenic Escherichia coli infection | 3 | 6.21 E-5 | 1.91 E-3 |
| Glycolysis Gluconeogenesis | 3 | 7.25 E-5 | 1.91 E-3 |
| Pancreatic cancer | 3 | 1.03 E-4 | 2.09 E-3 |
| Melanoma | 3 | 1.08 E-4 | 2.09 E-3 |
| Drug metabolism cytochrome P450 | 3 | 1.13 E-4 | 2.09 E-3 |

The effectiveness of the sparse constraint in the proposed method has been reasonably proved in the com-characteristic gene selection.

*2) Pathway Analysis:* To discover the mechanism of action of the discovered com-characteristic genes for the exploration of the connection between diseases, the public website KEGG was used to find the pathways. KEGG collects a series of chemical, genomic, and system functional data. The website is available at: http://www.kegg.jp/. First, the com-characteristic genes selected by the proposed method were sent to KEGG to compute the pathways. The summary of pathways found by the proposed method is listed in Table VI. The pathway graph with the highest *p*-value can be found in the supplementary file, where the genes in pink indicate human disease genes, the genes in blue indicate drug target genes, and the genes in green indicate human genes.

The pathways found by the proposed method were compared to the data of previously published articles. A large number of published articles reported the inseparable relationship between these pathways and cancer [69]–[75]. For example, it has been reported that the extracellular matrix (ECM) is composed of a complex mixture of structural and functional macromolecules. It also plays an important role in tissue and organ morphogenesis, e.g., in cell and tissue structure maintenance. Transmembrane molecules control the mechanism in cells and ECM. Their interaction reflects the influence of cellular biological activities. Mitra *et al.* [70] reported that the focal adhesion is a pivotal signaling component. Focal adhesion can regulate the structure of cell movement through a variety of molecular connections [70]. ECM and focal adhesion also indicate why cancer cells spread and transfer to other places very quickly. As shown in Table VI, many cancer pathways are still contained. Small cell lung cancer, bladder cancer, and PAAD have the highest genetic overlap. Several publications have shown that nonsmall cell lung cancer has a close relationship with PAAD [71], [72], and bladder cancer with CHOL and HNSC [73]–[76]. The relationship among these three diseases also indicates why these pathways can be found from the results of multiview data. Since these pathways are formed by the interaction of multiple genes, they indicate multiexclusive expressional levels.

### E. Results of Tumor Classification

Com-characteristic genes help to discover novel cancer markers and help to explore the potential connection between different diseases for early diagnosis and treatment. Since the proposed method introduces the supervised label information into the PCA model, its performance in classification is important. In this section, the proposed method has been used to classify the tumor samples. For comparison, $k$-nearest neighbor was used as baseline algorithm [77]. ACC denotes the percentage of correctly classified samples [78] and is defined as follows:

$$\text{ACC} = \frac{\sum_{i=1}^{n} \text{map}(p_i, q_i)}{n} \quad (17)$$

where $n$ represents the total number of samples, $p_i$ represents the prediction label, and $q_i$ represents the true label. $\text{map}(x, y)$ equals 1 if $x = y$; otherwise, $\text{map}(x, y) = 0$. A larger ACC indicates better classification performance. "Average ACC" denotes the average ACC value of different numbers of dimensions. The Average ACC is defined as follows:

$$\text{Average ACC} = \frac{\sum_{i=1}^{50} \text{ACC}_i}{50}. \quad (18)$$

The ACC results from 1 to 50 dimensions are recorded. Fig. 2 shows the changing curve of ACC with different dimensions. For each algorithm, the ACC results did not change much, while the dimension is large. PCA is related to the dimension of the data. For example, assuming that the dimension of the original data is $d$, then PCA can be used to obtain 1-$d$ features. LDA depends on the number of categories; however, it has no direct relationship with the dimensionality of the data. Assuming that the number of categories is $C$, then for LDA, its reduced dimensionality is generally $1 \sim C - 1$. Thus, in this experiment, the dimensions of LDA are 1–3 [41]. SLDA and SDA can achieve the ACC results of about 80% when the utilized dimension reduction is close to $C$. With the reduced dimensionality, significant information loss leads to poor results. PCA is an unsupervised dimensionality reduction method, which can be calculated without the knowledge of the sample label, while LDA is a supervised dimensionality reduction method that incorporates the label information. To provide a more intuitive classification results with the increase of the dimensions, the average ACC and variance from 1 to 50 dimensions are listed in Table VII. The results enable the following observations.

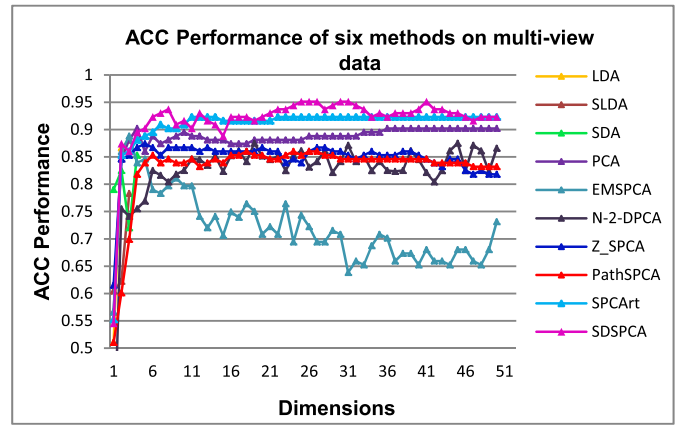

Fig. 2. Performance of all the tested methods on multiview data. The ACC results from 1 to 50 dimensions are presented.

| Methods | ACC±Variance |
|---------|--------------|
| LDA | 0.7524±0.0440 |
| SLDA | 0.6701±0.0097 |
| SDA | 0.7972±0.0033 |
| PCA | 0.8825±0.0022 |
| EMSPCA | 0.7201±0.0041 |
| N-2-DPCA | 0.8232±0.0067 |
| Z-SPCA | 0.8484±0.0013 |
| PathSPCA | 0.8305±0.0038 |
| SPCArt | 0.9085±0.0029 |
| SDSPCA | **0.9182±0.0033** |

1) When the dimension is small, the ACC values of all methods are poor. The loss of too much information is the reason for this.

2) With increasing the dimensionality, the value of ACC increases gradually and tends to stabilize after reaching a specific degree. This is due to the reduction of the information loss.

3) The proposed method is effective for the tumor classification compared to classical supervised methods and many sparse PCA methods.

4) PCA finds the direction of maximum variance, while LDA seeks the best discriminative direction. In bioinformatics, characteristic genes are easier to be detected by PCA due to their differential expression. Table VII shows that PCA-based methods achieve better ACC than LDA-based methods. The dimension reduction range of LDA is 1–3, and it incurs the largest fluctuation. For the high-dimensional gene expression data used in this experiment, the dimension is too large to significantly decrease. It is reasonable that PCA-based methods perform better than LDA-based methods in the average results of dimensions 1–50. SLDA and SDA obtain the sparse solution through LASSO and elastic net problems, respectively. Z-SPCA has the smallest fluctuation in different dimensions. EMSPA is far from satisfactory. Since SPCArt computes a sparse basis by rotating and truncating the basis of PCA, the performance of SPCArt is only secondary to SDSPCA. Its motivation differs from that of most other methods. The result of SPCArt is reasonable. In the tumor classification, several sparse PCA methods are less effective than PCA because sparsity causes data loss. However, SDSPCA outperforms the other methods and a similar model (N-2-DPCA). N-2-DPCA distinguishes different samples by their nuclear norm to describe the distance metric. Classifying samples based on the distance measurements are not as direct as supervised labels. Furthermore, N-2-DPCA does not consider the sparsity constraint, and PCs are still dense.

In summary, the introduction of the discriminative information in the proposed method could be the main reason for the prominent effect in the tumor classification. Therefore, the proposed method not only improves the interpretability of PCA (in the com-characteristic gene selection) but also overcomes the shortcomings of the high degree of ambiguity on the training samples for unsupervised learning (in the tumor classification).

## IV. Conclusion

This paper presents the novel method SDSPCA that incorporates both supervised discriminative information and sparse constraint into a PCA model. SDSPCA not only improves the interpretability of PCA but also overcomes the shortcoming of high-degree class ambiguity in unsupervised learning. SDSPCA shows the competitive capability in the com-characteristic gene selection and the tumor classification on multiview biological data. The identified com-characteristic genes provide an unprecedented opportunity to explore the potential connection among different diseases on multiview data. Furthermore, the good tumor classification results confirmed the capability of the proposed method for biologic data.

There are, nevertheless, limitations of the proposed method. For example, the proposed SDSPCA, as a linear dimensionality reduction method, cannot capture the nonlinear geometric structure of data. Moreover, the method is sensitive to outliers due to the square calculations in the loss function. It is, therefore, necessary to further enhance this method to improve its robustness. In the future work, the proposed method will be further improved by more practical biological applications.

## References

[1] R. A. Irizarry, Z. Wu, and H. A. Jaffee, "Comparison of affymetrix GeneChip expression measures," *Bioinformatics*, vol. 22, no. 7, pp. 789–794, 2006.

[2] L. Liu, X. Wang, and H. Chen, "DNA microarray and gene expression," *Progr. Veterinary Med.*, vol. 24, 2003.

[3] D. Edsgärd, P. Johnsson, and R. Sandberg, "Identification of spatial expression trends in single-cell gene expression data," *Nature Methods*, vol. 15, pp. 339–342, Mar. 2018.

[4] M. S. Dona, L. A. Prendergast, S. Mathivanan, S. Keerthikumar, and A. Salim, "Powerful differential expression analysis incorporating network topology for next-generation sequencing data," *Bioinformatics*, vol. 33, p. 1505, 2017.

[5] J. Song and J. D. Hart, "Bootstrapping in a high dimensional but very low-sample size problem," *J. Stat. Comput., Simul.*, vol. 80, no. 8, pp. 825–840, 2010.

[6] A. Serra, M. Fratello, V. Fortino, G. Raiconi, R. Tagliaferri, and D. Greco, "MVDA: A multi-view genomic data integration methodology," *BMC Bioinf.*, vol. 16, p. 261, Aug. 2015.

[7] Y. Li, F. X. Wu, and A. Ngom, "A review on machine learning principles for multi-view biological data integration," *Brief Bioinf.*, vol. 19, no. 2, pp. 325–340, 2018.

[8] J. X. Liu, Y. Xu, C. H. Zheng, H. Kong, and Z. H. Lai, "RPCA-based tumor classification using gene expression data," *IEEE/ACM Trans. Comput. Biol. Bioinf.*, vol. 12, no. 4, pp. 964–970, Jul. 2015.

[9] D. V. Nguyen and D. M. Rocke, "Tumor classification by partial least squares using microarray gene expression data," *Bioinformatics*, vol. 18, no. 1, pp. 39–50, 2002.

[10] B. Q. Huynh, H. Li, and M. Giger, "Digital mammographic tumor classification using transfer learning from deep convolutional neural networks," *J. Med. Imag.*, vol. 3, no. 3, p. 034501, Aug. 2016.

[11] S. Xiang, F. Nie, G. Meng, and C. Pan, "Discriminative least squares regression for multiclass classification and feature selection," *IEEE Trans. Neural Netw. Learn. Syst.*, vol. 23, no. 11, pp. 1738–1754, Nov. 2012.

[12] E. Adeli *et al.*, "Semi-supervised discriminative classification robust to sample-outliers and feature-noises," *IEEE Trans. Pattern Anal. Mach. Intell.*, vol. 41, no. 2, pp. 515–522, Feb. 2018.

[13] M. Xiong, W. Li, J. Zhao, L. Jin, and E. Boerwinkle, "Feature (Gene) selection in gene expression-based tumor classification," *Mol. Genet. Metabolism*, vol. 73, no. 3, pp. 239–247, 2001.

[14] Q. Shen, W.-M. Shi, and W. Kong, "Hybrid particle swarm optimization and tabu search approach for selecting genes for tumor classification using gene expression data," *Comput. Biol. Chem.*, vol. 32, no. 1, pp. 53–60, 2008.

[15] H.-I. Suk, S.-W. Lee, and D. Shen, "Deep sparse multi-task learning for feature selection in Alzheimer's disease diagnosis," *Brain Struct. Function*, vol. 221, no. 5, pp. 2569–2587, 2016.

[16] J.-X. Liu, Y.-T. Wang, C.-H. Zheng, W. Sha, J.-X. Mi, and Y. Xu, "Robust PCA based method for discovering differentially expressed genes," *BMC Bioinform.*, vol. 14, no. 8, pp. 1–10, 2013.

[17] I. T. Jolliffe, *Principal Component Analysis*, vol. 87. Berlin, Germany: Springer, 1986, pp. 41–64.

[18] J.-X. Liu, Y. Xu, Y.-L. Gao, C.-H. Zheng, D. Wang, and Q. Zhu, "A class-information-based sparse component analysis method to identify differentially expressed genes on RNA-seq data," *IEEE/ACM Trans. Comput. Biol. Bioinf.*, vol. 13, no. 2, pp. 392–398, Mar./Apr. 2016.

[19] Z. Lai, Y. Xu, Q. Chen, J. Yang, and D. Zhang, "Multilinear sparse principal component analysis," *IEEE Trans. Neural Netw. Learn. Syst.*, vol. 25, no. 10, pp. 1942–1950, Oct. 2014.

[20] R. Guo, M. Ahn, and H. Zhu, "Spatially weighted principal component analysis for imaging classification," *J. Comput. Graph. Statist.*, vol. 24, no. 1, pp. 274–296, 2015.

[21] L. I. Kuncheva and W. J. Faithfull, "PCA feature extraction for change detection in multidimensional unlabeled data," *IEEE Trans. Neural Netw. Learn. Syst.*, vol. 25, no. 1, pp. 69–80, Jan. 2014.

[22] R. Zhang, J. Tao, R. Lu, and Q. Jin, "Decoupled ARX and RBF neural network modeling using PCA and GA optimization for nonlinear distributed parameter systems," *IEEE Trans. Neural Netw. Learn. Syst.*, vol. 29, no. 2, pp. 457–469, Feb. 2018.

[23] L. Qi, W. Dou, and J. Chen, "Weighted principal component analysis-based service selection method for multimedia services in cloud," *Computing*, vol. 98, nos. 1–2, pp. 195–214, 2016.

[24] N. Duforet-Frebourg, K. Luu, G. Laval, E. Bazin, and M. G. Blum, "Detecting genomic signatures of natural selection with principal component analysis: Application to the 1000 Genomes data," *Mol. Biol. Evol.*, vol. 33, no. 4, pp. 1082–1093, 2016.

[25] Y. K. Yeong and R. K. Kapania, "Neural networks for inverse problems using principal component analysis and orthogonal arrays," *AIAA J.*, vol. 44, no. 7, pp. 1628–1634, 2015.

[26] A. d'Aspremont, L. El Ghaoui, M. I. Jordan, and G. R. G. Lanckriet, "A direct formulation for sparse PCA using semidefinite programming," *Siam Rev.*, vol. 49, no. 3, pp. 434–448, 2006.

[27] D. Meng, Q. Zhao, and Z. Xu, "Improve robustness of sparse PCA by $L_1$-norm maximization," *Pattern Recognit.*, vol. 45, no. 1, pp. 487–497, Jan. 2012.

[28] J. K. Siddiqui *et al.*, "IntLIM: Integration using linear models of metabolomics and gene expression data," *BMC Bioinf.*, vol. 19, p. 81, Mar. 2018.

[29] X. Gao, X. Gao, X. Gao, and X. Gao, "Large sparse cone non-negative matrix factorization for image annotation," *ACM Trans. Intell. Syst. Technol.*, vol. 8, no. 3, p. 37, 2017.

[30] D. Tao, J. Cheng, M. Song, and X. Lin, "Manifold ranking-based matrix factorization for saliency detection," *IEEE Trans. Neural Netw. Learn. Syst.*, vol. 27, no. 6, pp. 1122–1134, Jun. 2016.

[31] D. Tao, J. Cheng, X. Gao, X. Li, and C. Deng, "Robust sparse coding for mobile image labeling on the cloud," *IEEE Trans. Circuits Syst. Video Technol.*, vol. 27, no. 1, pp. 62–72, Jan. 2017.

[32] H. Zou, T. Hastie, and R. Tibshirani, "Sparse principal component analysis," *J. Comput. Graph. Statist.*, vol. 15, no. 2, pp. 265–286, 2006.

[33] C. D. Sigg and J. M. Buhmann, "Expectation-maximization for sparse and non-negative PCA," in *Proc. Int. Conf.*, 2008, pp. 960–967.

[34] A. d'Aspremont, F. Bach, and L. E. Ghaoui, "Optimal solutions for sparse principal component analysis," *J. Mach. Learn. Res.*, vol. 9, pp. 1269–1294, Jul. 2007.

[35] H. Shen and J. Z. Huang, "Sparse principal component analysis via regularized low rank matrix approximation," *J. Multivariate Anal.*, vol. 99, no. 6, pp. 1015–1034, Jul. 2008.

[36] M. Journée, Y. Nesterov, P. Richtárik, and R. Sepulchre, "Generalized power method for sparse principal component analysis," *J. Mach. Learn. Res.*, vol. 11, pp. 517–553, Mar. 2010.

[37] Z. Lu and Y. Zhang, "An augmented Lagrangian approach for sparse principal component analysis," *Math. Program.*, vol. 135, nos. 1–2, pp. 149–193, 2012.

[38] B. Moghaddam, Y. Weiss, and S. Avidan, "Spectral bounds for sparse PCA: Exact and greedy algorithms," in *Proc. Adv. Neural Inf. Process. Syst.*, 2005, pp. 915–922.

[39] D. Meng, H. Cui, Z. Xu, and K. Jing, "Following the entire solution path of sparse principal component analysis by coordinate-pairwise algorithm," *Data Knowl. Eng.*, vol. 88, pp. 25–36, Nov. 2013.

[40] Z. Hu, G. Pan, Y. Wang, and Z. Wu, "Sparse principal component analysis via rotation and truncation," *IEEE Trans. Neural Netw. Learn. Syst.*, vol. 27, no. 4, pp. 875–890, Apr. 2016.

[41] X. Yong, J. Y. Yang, and J. Zhong, "Theory analysis on FSLDA and ULDA," *Pattern Recognit.*, vol. 36, no. 12, pp. 3031–3033, 2003.

[42] Z. Feng, M. Yang, L. Zhang, Y. Liu, and D. Zhang, "Joint discriminative dimensionality reduction and dictionary learning for face recognition," *Pattern Recognit.*, vol. 46, no. 8, pp. 2134–2143, Aug. 2013.

[43] L. Clemmensen, T. Hastie, D. Witten, and B. Ersboll, "Sparse discriminant analysis," *Technometrics*, vol. 53, no. 4, pp. 406–413, 2011.

[44] Z. Jiang, Z. Lin, and L. S. Davis, "Label consistent K-SVD: Learning a discriminative dictionary for recognition," *IEEE Trans. Pattern Anal. Mach. Intell.*, vol. 35, no. 11, pp. 2651–2664, Nov. 2013.

[45] M. Yang, L. Zhang, X. Feng, and D. Zhang, "Sparse representation based Fisher discrimination dictionary learning for image classification," *Int. J. Comput. Vis.*, vol. 109, no. 3, pp. 209–232, Sep. 2014.

[46] Z. Qiao, L. Zhou, and J. Z. Huang, "Sparse linear discriminant analysis with applications to high dimensional low sample size data," *IAENG Int. J. Appl. Math.*, vol. 39, no. 1, pp. 48–60, Jan. 2009.

[47] Z. Lai, Y. Xu, J. Yang, J. Tang, and D. Zhang, "Sparse tensor discriminant analysis," *IEEE Trans. Image Process.*, vol. 22, no. 10, pp. 3904–3915, Oct. 2013.

[48] F. Zhang, J. Yang, J. Qian, and Y. Xu, "Nuclear norm-based 2-DPCA for extracting features from images," *IEEE Trans. Neural Netw. Learn. Syst.*, vol. 26, no. 10, pp. 2247–2260, Oct. 2015.

[49] X. Zhu, X. Li, S. Zhang, Z. Xu, L. Yu, and C. Wang, "Graph PCA hashing for similarity search," *IEEE Trans. Multimedia*, vol. 19, no. 9, pp. 2033–2044, Sep. 2017.

[50] C. M. Feng, Y. L. Gao, J. X. Liu, J. Wang, D. Q. Wang, and C. G. Wen, "Joint $L_{1/2}$-norm constraint and graph-Laplacian PCA method for feature extraction," *BioMed Res. Int.*, vol. 2017, pp. 1–14, Mar. 2017.

[51] C.-M. Feng, Y.-L. Gao, J.-X. Liu, C.-H. Zheng, and J. Yu, "PCA based on graph Laplacian regularization and P-norm for gene selection and clustering," *IEEE Trans. Nanobiosci.*, vol. 16, no. 4, pp. 257–265, Jun. 2017.

[52] Q. Zhang and B. Li, "Discriminative K-SVD for dictionary learning in face recognition," in *Proc. Comput. Vis. Pattern Recognit.*, 2010, pp. 2691–2698.

[53] E. Barshan, A. Ghodsi, Z. Azimifar, and M. Z. Jahromi, "Supervised principal component analysis: Visualization, classification and regression on subspaces and submanifolds," *Pattern Recognit.*, vol. 44, no. 7, pp. 1357–1371, 2011.

[54] B. Jiang, C. Ding, and J. Tang, "Graph-Laplacian PCA: Closed-form solution and robustness," in *Proc. IEEE Conf. Comput. Vis. Pattern Recognit.*, 2013, pp. 3492–3498.

[55] F. Nie, H. Huang, X. Cai, and C. H. Q. Ding, "Efficient and robust feature selection via joint $L_{2,1}$-norms minimization," in *Proc. Adv. Neural Inf. Process. Syst.*, 2010, pp. 1813–1821.

[56] C. Ding, D. Zhou, X. He, and H. Zha, "R1-PCA: Rotational invariant $L_1$-norm principal component analysis for robust subspace factorization," in *Proc. Int. Conf. Mach. Learn.*, 2006, pp. 281–288.

[57] S. Yang, C. Hou, F. Nie, and Y. Wu, "Unsupervised maximum margin feature selection via $L_{2,1}$-norm minimization," *Neural Comput. Appl.*, vol. 21, no. 7, pp. 1791–1799, 2012.

[58] C. Hou, F. Nie, D. Yi, and Y. Wu, "Feature selection via joint embedding learning and sparse regression," in *Proc. Int. Joint Conf. Artif. Intell. (IJCAI)*, Barcelona, Spain, Jul. 2011, pp. 1324–1329.

[59] L. S. Weinstein, J. Liu, A. Sakamoto, T. Xie, and M. Chen, "Minireview: GNAS: Normal and abnormal functions," *Endocrinology*, vol. 145, no. 12, pp. 5459–5464, 2005.

[60] S. Weissmueller, M. Saborowski, E. Manchado, V. Thapar, and S. W. Lowe, "Abstract C54: Pdgfrb is an essential mediator of p53 (mut)-driven metastasis in pancreatic cancer," *Cancer Res.*, vol. 73, p. C54, Oct. 2014.

[61] M. Munoz, M. Henderson, M. Haber, and M. Norris, "Role of the MRP1/ABCC1 multidrug transporter protein in cancer," *Int. Union Biochem., Molecular Biol. Life*, vol. 59, no. 12, pp. 752–757, 2008.

[62] U. Srimunta *et al.*, "High expression of ABCC1 indicates poor prognosis in intrahepatic cholangiocarcinoma," *Asian Pacific J. Cancer Prevention*, vol. 13, pp. 125–130, Jan. 2012.

[63] S. K. Srivastava *et al.*, "MicroRNA-150 directly targets MUC4 and suppresses growth and malignant behavior of pancreatic cancer cells," *Carcinogenesis*, vol. 32, no. 12, pp. 1832–1839, 2011.

[64] M. A. Macha *et al.*, "MUC4 regulates cellular senescence in head and neck squamous cell carcinoma through p16/Rb pathway," *Oncogene*, vol. 34, pp. 1698–1708, Jan. 2015.

[65] H. Shibahara *et al.*, "MUC4 is a novel prognostic factor of intrahepatic cholangiocarcinoma-mass forming type," *Hepatology*, vol. 39, no. 1, pp. 220–229, 2004.

[66] J. C. Ibbitt, "FGFR4 overexpression in pancreatic cancer is mediated by an intronic enhancer activated by HNF1alpha," *Oncogene*, vol. 21, no. 54, pp. 8251–8261, 2002.

[67] S. Streit, J. Bange, A. Fichtner, S. Ihrler, W. Issing, and A. Ullrich, "Involvement of the FGFR4 Arg388 allele in head and neck squamous cell carcinoma," *Int. J. Cancer*, vol. 111, no. 2, pp. 213–217, 2004.

[68] A. Kapoor *et al.*, "Yap1 activation enables bypass of oncogenic Kras addiction in pancreatic cancer," *Cell*, vol. 158, no. 1, pp. 97–185, 2014.

[69] T. Ahrens *et al.*, "Interaction of the extracellular matrix (ECM) component Hyaluronan (HA) with its principle cell surface receptor CD44 augments human melanoma cell proliferation," *J. Dermatol. Sci.*, vol. 16, p. S121, 1998.

[70] S. K. Mitra, D. A. Hanson, and D. D. Schlaepfer, "Focal adhesion kinase: In command and control of cell motility," *Nature Rev. Mol. Cell Biol.*, vol. 6, no. 1, pp. 56–68, 2005.

[71] S. Noble and K. L. Goa, "Gemcitabine. A review of its pharmacology and clinical potential in non-small cell lung cancer and pancreatic cancer," *Drugs*, vol. 54, no. 3, pp. 447–472, 1997.

[72] Z. Meng *et al.*, "Pilot study of Huachansu in patients with hepatocellular carcinoma, non-small cell lung cancer, or pancreatic cancer," *Cancer*, vol. 115, no. 22, pp. 5309–5318, 2009.

[73] P. Zhou *et al.*, "The epithelial to mesenchymal transition (EMT) and cancer stem cells: Implication for treatment resistance in pancreatic cancer," *Mol. Cancer*, vol. 16, no. 1, pp. 52-1–52-11, 2017.

[74] T. Cermik *et al.*, "Evolving role of FDG-PET imaging in the management of patients with gall bladder cancer and cholangiocarcinoma," *J. Nucl. Med.*, vol. 47, no. 1, p. 1715, 2006.

[75] K. Biggers and N. Scheinfeld, "VB4-845, a conjugated recombinant antibody and immunotoxin for head and neck cancer and bladder cancer," *Current Opinion Mol. Therapeutics*, vol. 10, no. 2, pp. 176–186, 2008.

[76] F. Schedel *et al.*, "mTOR inhibitors show promising *in vitro* activity in bladder cancer and head and neck squamous cell carcinoma," *Oncol. Rep.*, vol. 25, no. 3, pp. 763–768, 2011.

[77] M.-L. Zhang and Z.-H. Zhou, "ML-KNN: A lazy learning approach to multi-label learning," *Pattern Recognit.*, vol. 40, no. 7, pp. 2038–2048, 2007.

[78] I. Guyon, J. Weston, S. Barnhill, and V. Vapnik, "Gene selection for cancer classification using support vector machines," *Mach. Learn.*, vol. 46, nos. 1–3, pp. 389–422, 2002.

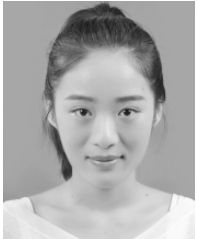

**Chun-Mei Feng** received the M.S. degree from Qufu Normal University, Rizhao, China, in 2018. She is currently pursuing the Ph.D. degree with the School of Information Science and Technology, Harbin Institute of Technology, Shenzhen, China.

Her current research interests include deep learning, feature selection, pattern recognition, and bioinformatics.

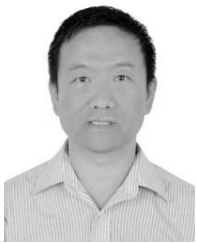

**Yong Xu** (M'06–SM'15) received the B.S. and M.S. degrees from the Air Force Institute of Meteorology, Nanjing, China, in 1994 and 1997, respectively, and the Ph.D. degree in pattern recognition and intelligence system from the Nanjing University of Science and Technology, Nanjing, China, in 2005.

From 2005 to 2007, he was a Post-Doctoral Research Fellow with the Harbin Institute of Technology, Shenzhen, China, where he is currently a Professor. His current interests include pattern recognition, machine learning, and bioinformatics.

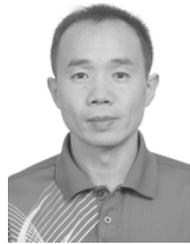

**Jin-Xing Liu** (M'12) received the B.S. degree in electronic information and electrical engineering from Shandong University, Jinan, China, in 1997, the M.S. degree in control theory and control engineering from Qufu Normal University, Rizhao, China, in 2003, and the Ph.D. degree in computer simulation and control from the South China University of Technology, Guangzhou, China, in 2008.

He is currently an Associate Professor with the School of Information Science and Engineering, Qufu Normal University. His current research interests include pattern recognition, machine learning, and bioinformatics.

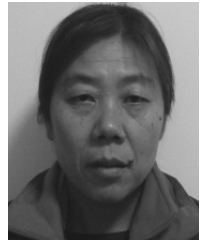

**Ying-Lian Gao** received the B.S. and M.S. degrees from Qufu Normal University, Rizhao, China, in 1997 and 2000, respectively.

She is currently with the Library of Qufu Normal University, Qufu Normal University. Her current interests include data mining and pattern recognition.

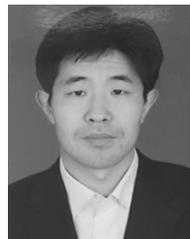

**Chun-Hou Zheng** (M'10) received the B.Sc. degree in physics education and the M.Sc. degree in control theory and control engineering from Qufu Normal University, Rizhao, China, in 1995 and 2001, respectively, and the Ph.D. degree in pattern recognition and intelligent system from the University of Science and Technology of China, Hefei, China, in 2006.

From 2007 to 2009, he was a Post-Doctoral Fellow with the Hefei Institutes of Physical Science, Chinese Academy of Sciences, Hefei. From 2009 to 2010, he was a Post-Doctoral Fellow with the Department of Computing, The Hong Kong Polytechnic University, Hong Kong. He is currently a Professor with the School of Computer Science and Technology, Anhui University, Hefei. His research interests include pattern recognition and bioinformatics.